\documentclass{article}

\usepackage{arxiv}

\usepackage[utf8]{inputenc} 
\usepackage[T1]{fontenc}    
\usepackage{hyperref}       
\usepackage{url}            
\usepackage{booktabs}       
\usepackage{amsfonts}       
\usepackage{nicefrac}       
\usepackage{microtype}      
\usepackage{lipsum}
\usepackage{graphicx}
\graphicspath{ {./images/} }

\usepackage{microtype}
\usepackage{graphicx}
\usepackage{subfigure}
\usepackage{booktabs} 
\usepackage{breqn}

\usepackage[round]{natbib}

\newcommand{\R}{\mathbb{R}}

\usepackage{float}
\usepackage{graphicx}
\usepackage{lipsum}
\usepackage[utf8]{inputenc} 
\usepackage[T1]{fontenc}    
\usepackage{hyperref}       
\usepackage{url}            
\usepackage{booktabs}       
\usepackage{amsfonts}       
\usepackage{nicefrac}       
\usepackage{microtype}      

\usepackage{amsmath}

\usepackage[ruled,vlined]{algorithm2e}

\title{Distributional Gaussian Process Layers for Outlier Detection in Image Segmentation}

\author{
 Sebastian G. Popescu \\
  Imperial College London\\
  \texttt{s.popescu16@ic.ac.uk} \\
   \And
  David J. Sharp \\
  Imperial College London \\
  \texttt{david.sharp@ic.ac.uk} \\
  \And
  James H. Cole \\
  University College London \\
  \texttt{james.cole@ucl.ic.ac.uk} \\
  \AND
  Konstantinos Kamnitsas \\
  Imperial College London \\
  \texttt{konstantinos.kamnitsas12@ic.ac.uk} \\
  \And
  Ben Glocker \\
  Imperial College London \\
  \texttt{b.glocker@ic.ac.uk} \\
}

\begin{document}
\maketitle
\begin{abstract}

    We propose a parameter efficient Bayesian layer for hierarchical convolutional Gaussian Processes that incorporates Gaussian Processes operating in Wasserstein-2 space to reliably propagate uncertainty. This directly replaces convolving Gaussian Processes with a distance-preserving affine operator on distributions. Our experiments on brain tissue-segmentation show that the resulting architecture approaches the performance of well-established deterministic segmentation algorithms (U-Net), which has never been achieved with previous hierarchical Gaussian Processes. Moreover, by applying the same segmentation model to out-of-distribution data (i.e., images with pathology such as brain tumors), we show that our uncertainty estimates result in out-of-distribution detection that outperforms the capabilities of previous Bayesian networks and reconstruction-based approaches that learn normative distributions.

\end{abstract}

\section{Introduction}

Deep learning methods have achieved state-of-the-art results on a plethora of medical image segmentation tasks \cite{zhou2020review}. However, their application in clinical settings is very limited due to issues pertaining to lack of reliability and miscalibration of estimated confidence in predictions. Most research into incorporating uncertainty into medical image segmentation has gravitated around modelling inter-rater variability and the inherent aleatoric uncertainty associated to the dataset, which can be caused by noise or inter-class ambiguities. However, not much focus has been placed on how models behave when processing unexpected input, which differ from what has been processed during training, often called anomalies, outliers or out-of-distribution samples.

Out-of-distribution detection (OOD) in medical imaging has been mostly approached through the lens of reconstruction-based techniques involving some form of encoder-decoder network trained on normative datasets \cite{chen2019unsupervised}. Conversely, we focus on enhancing task-specific models (e.g. a segmentation model) with reliable uncertainty quantification that enables outlier detection. Standard deep neural networks (DNNs), despite their high predictive performance, often show poor calibration between confidence estimates and prediction accuracy when processing unseen samples that are not from the data manifold of training set (e.g. in the presence of pathology). To alleviate this, Bayesian approaches that assign posteriors over both weights and function space have been proposed \cite{wilson2020bayesian}. In this paper we follow an alternative approach, using Gaussian Processes (GP) as the building block for deep Bayesian networks. The usage of GP for image classification has garnered interest in the past years. Convolutional GP were stacked on feed forward GP layers applied in a convolutional manner, with promising improvements in accuracy compared to their shallow counterpart \cite{blomqvist2018deep}. We expand on the latter work, by introducing a simpler convolutional mechanism, which does not require convolving GP at each layer and hence alleviates the computational cost of optimizing over inducing points' locations residing in high dimensional spaces. We propose a plug-in Bayesian layer more amenable to CNN architectures, which replaces the convolved filter followed by parametric activation function with a distance-preserving affine operator on stochastic layers for convolving the Gaussian measures from the previous layer of a hierarchical GP, and subsequently using Distributional Gaussian Processes (DistGP) \cite{popescu2020hierarchical} as a one-to-one mapping, essentially acting as a non-parametric activation function. DistGP were shown to be better at propagating outliers, as given by high variance, compared to standard GP due to their kernel design.

\subsection{Related work}

Research into Bayesian models has focused on a separation of uncertainty into two different types, aleatoric (data intrinsic) and epistemic (model parameter uncertainty). The former is irreducible, given by noise in the data acquisition process and has been extensively used for medical image segmentation \cite{monteiro2020stochastic}, whereas the latter can be reduced by giving the model more data. It has also found itself used in segmentation tasks \cite{nair2020exploring}. However, none of these works test how their models behave in the presence of outliers. Another type of uncertainty is introduced in \cite{popescu2020hierarchical}, where Sparse Gaussian Processes (SGP) \cite{hensman2013gaussian} are decomposed into components that separate within-data manifold uncertainty from distributional uncertainty. The latter increases away from the training data manifold, which we use here as a measure of OOD. To find similar metrics of OOD we explored general OOD literature for models which we can adapt for image segmentation. Variations of classical one-versus-all models have been adapted to neural networks \cite{padhy2020revisiting,franchi2020one}. The closest work that we could find to our proposed approach uses a deep network as a feature extractor for an RBF network \cite{van2020simple}. The authors quantify epistemic uncertainty as the L2 distance between a given data point and centroids corresponding to different classes, much alike the RBF kernels and the inducing point approach seen in SGP.

\subsection{Contributions}

This work makes the following main contributions:
 \begin{itemize}
  \item We introduce a Bayesian layer that combines convolved affine operators that are upper bounded in Wasserstein-2 space and DistGP as ``activation functions'', which results in an expressive non-parametric convolutional layer with Lipschitz continuity and reliable uncertainty quantification. 
  \item We show for the first time that a GP-based convolutional architecture achieves competitive results in segmentation tasks in comparison to a U-Net.
  \item We demonstrate improved OOD results compared to Bayesian models and reconstruction-based models.
\end{itemize}

\section{Hierarchical GP with Wasserstein-2 kernels}

We denote input points $X=\left( x_{1},...,x_{n} \right)$ and the output vector $Y = \left( y_{1},...,y_{n} \right)$. We consider a Deep GP (DGP), which is a composition of functions $p_{L} = p_{L} \circ ... \circ p_{1}$. Each $p_{l}$ is given by a $GP(m,k)$ prior on the stochastic function $F_{l}$, where under standard Gaussian identities we have: 
\begin{gather}
    p(Y|F_{L})= \mathcal{N}(Y|F_{L},\beta) \\
    p(F_{l}|U_{l};F_{l-1},Z_{l-1}) = \mathcal{N}(F_{l}|K_{nm}K_{mm}^{-1}U_{l}, K_{nn} - K_{nm}K_{mm}^{-1}K_{mn};X,Z) \\
    p(U_{l};Z_{l}) = \mathcal{N}(U_{l}|0,K_{mm})
\end{gather}
where $Z_{l}$ and $U_{l}$ are the locations and values respectively of the GP's inducing points. $\beta$ represents the likelihood noise and $K_{\cdot,\cdot}$ represents the kernel. A DGP is then defined as a stack of shallow SGP operating in Euclidean space with the prior being:
\begin{equation}
p(Y) = \underbrace{p(Y|F_{L})}_{\text{likelihood}}\underbrace{\prod_{l=1}^{L} p(F_{l}|U_{l};F_{l-1},Z_{l-1})p(U_{l})}_{\text{Euclidean prior}}
\end{equation}

where for brevity of notation we denote $Z_{0}\!=\!X$. Differently from DGP, in a Hierarchical DistGP \cite{popescu2020hierarchical} all layers except the last one are deterministic operations on Gaussian measures. Concretely, it has the following joint density prior:

\begin{equation}
\begin{aligned}
p(Y,\{F_{l},U_{l}\}_{l=1}^{L}) = \underbrace{p(Y|F_{L})}_{likelihood}\underbrace{\prod_{l=2}^{L} p(F_{l}|F_{l-1},U_{l};Z_{l-1})p(U_{l})}_{ \text{Wasserstein space prior}}\underbrace{ p(F_{1}|U_{1};X)p(U_{1})}_{\text{Euclidean space prior}}
\end{aligned}
\end{equation}

A factorized posterior between layers and dimensions is introduced $q(F_{L},\{U_{l}\}_{l=1}^{L}) =  p(F_{L}|U_{L};Z_{L-1})\prod_{l=1}^{L}q(U_{l})$, where for $1\leq l \leq L$ the approximate posterior over is $U_{l} \sim \mathbb{N}(m_{l},\Sigma_{l})$ and $Z_{l} \sim \mathbb{N}(z_{m_{l}},Z_{\Sigma_{l}})$. $Z_{0}$ is optimized in standard Euclidean space. Using Jensen's inequality we arrive at the evidence lower bound (ELBO):
\begin{equation}
L = \mathbb{E}_{q(F_{L},\{U_{l}\}_{l=1}^{L})} \frac{p(Y,F_{L},\{U_{l}\}_{l=1}^{L})}{q(F_{L},\{U_{l}\}_{l=1}^{L})}  = \mathbb{E}_{q(F_{L},\{U_{l}\}_{l=1}^{L})} p(Y|F_{L}) - \sum_ {l=1}^{L}KL(q(U_{l})|p(U_{l}))
\end{equation}

\section{Convolutionally Warped DistGP \& Activation Function}

\begin{figure}[!htb]
\centering
  \includegraphics[width=0.7\linewidth]{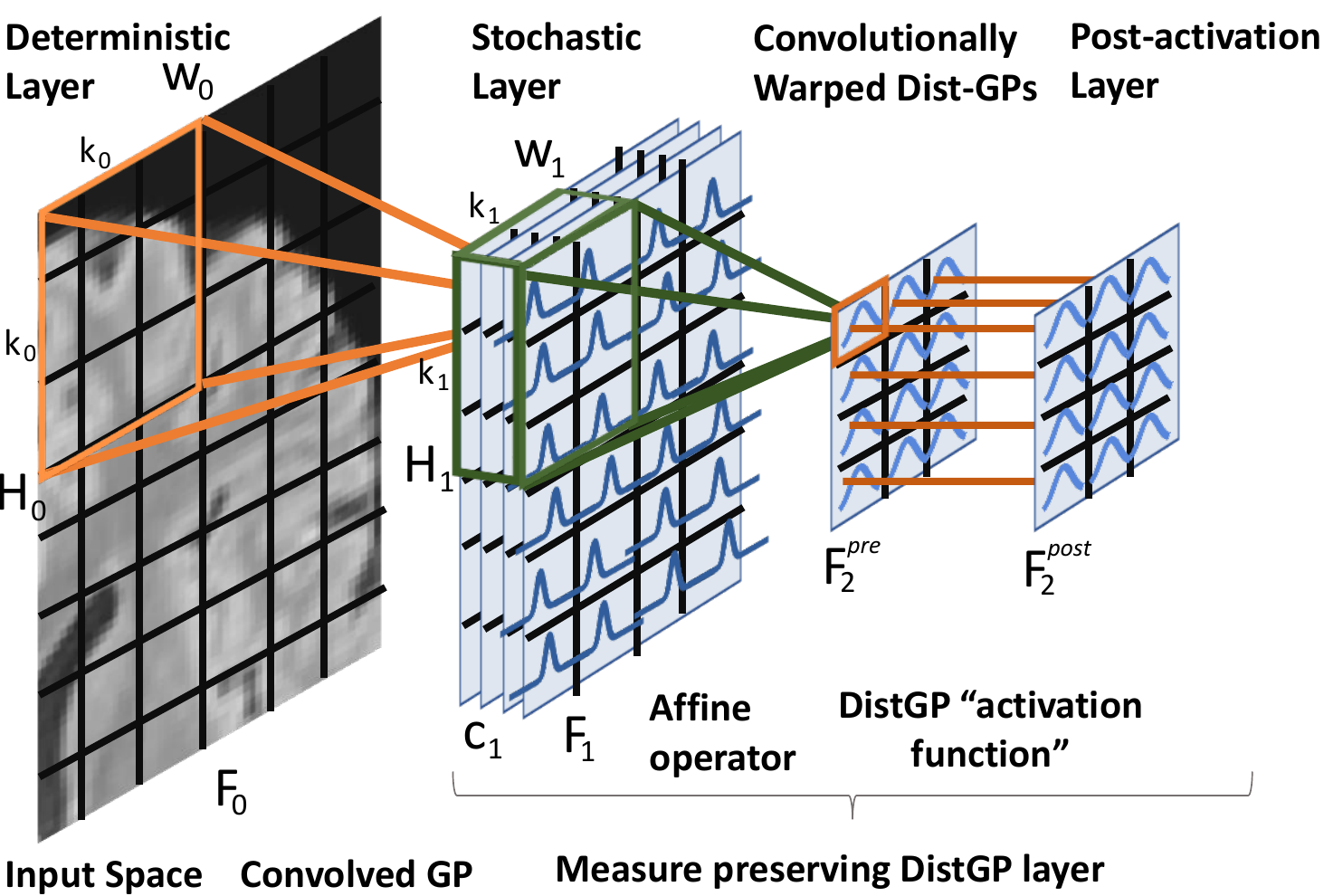}
  \caption{Schematic of measure-preserving DistGP layer.}
  \label{fig:schematic_algorithm}
\end{figure}

For ease of notation and graphical representation we describe the case of the input being a 2D image, with no loss of generality. We denote the image's representation $F_{l} \in \R^{H_{l},W_{l},C_{l}}$ with width $W_{l}$, height $H_{l}$ and $C_{l}$ channels at the l-th layer of a multi-layer model. $F_{0}$ is the image.
Consider a square kernel of size $k_{l}\!\times\!k_{l}$. We denote with $F^{[p,k_{l}]}_{l} \in \R^{k_{l},k_{l},C_{l}}$ the $p$-th patch of $F_l$, which is the area of $F_l$ that the kernel covers when overlaid at position $p$ during convolution (e.g. orange square for a $3\!\times\!3$ kernel in Figure \ref{fig:schematic_algorithm}).
We introduce the convolved $GP_{0} : F^{[p,k_{0}]}_{0} \rightarrow \mathcal{N}(m,k)$ with $Z_{0} \in \R^{k_{0},k_{0},C_{0}}$ to be the SGP operating on the Euclidean space of patches of the input image in a similar fashion to the layers introduced in \cite{blomqvist2018deep}. For $ 1 \leq l \leq L $ we introduce affine embeddings
$A_{l} \in \R^{k_{l},k_{l},C_{l-1},C_{l,pre}}$, where $C_{l,pre}$ denotes the number of channels in the \emph{pre-activation} (e.g. $F_{2}^{pre}$ in Figure \ref{fig:schematic_algorithm}), which are convolved on the previous stochastic layer in the following manner:

\begin{gather}
    m(F^{pre}_{l}) = Conv_{2D}(m(F_{l-1}),A_{l}) \label{eq:conv_mu} \\
    var(F^{pre}_{l}) = Conv_{2D}(var(F_{l-1}),A_{l}^{2}) \label{eq:conv_var}
\end{gather}

The affine operator is sequentially applied on the mean, respectively variance components of the previous layer $ F_{l-1}$ so as to propagate the Gaussian measures to the next pre-activation layer $ F^{pre}_{l}$. To obtain the post-activation layer, we apply a $DistGP_{l} : F^{pre,[p,1]}_{l} \rightarrow \mathcal{N}(m,k)$ in a many-to-one manner on the pre-activation patches to arrive at $F^{post}_{l}$. Figure \ref{fig:schematic_algorithm} depicts this new module, entitled "Measure preserving DistGP" layer. In \cite{blomqvist2018deep} the convolved GP is used across the entire hierarchy, thereby inducing points are in high-dimensional space ($k_{l}^{2}*C_{l}$). In our case, the convolutional process is replaced by an inducing points free affine operator, with inducing points in low-dimensional space ($C_{l,pre}$) for the DistGP activation functions. The affine operator outputs $C_{l,pre}$, which is taken to be higher than the associated output space of DistGP activation functions $C_{l}$. Hence, the affine operator can cheaply expand the channels, in constrast to the layers in \cite{blomqvist2018deep} which would require high-dimensional multi-output GP. We motivate the preservation of distance in Wasserstein-2 space in the following section.


\section{Imposing Lipschitz Conditions in Convolutionally Warped DistGP}
\label{sec:conv_dist_gp}

If a sample is identified as an outlier at certain layer, respectively being flagged with high variance, in an ideal scenario we would like to preserve that status throughout the remainder of the network. As the kernels operate in Wasserstein-2 space, the distance of a data point's first two moments with respect to inducing points is vital. Hence, we would like our network to vary smoothly between layers, so that similar objects in previous layers get mapped into similar spaces in the Wasserstein-2 domain. In this section, we accomplish this by quantifying the \textit{"Lipschitzness"} of our "Measure preserving DistGP" layer and by imposing constraints on the affine operators so that they preserve distances in Wasserstein-2 space.

\paragraph{\textbf{Definition}} We define the Wasserstein-2 distance as
$W_{2}(\mu,\nu) = (\inf_{\pi\in\Pi(\mu,\nu)} \int[x-y]^{2}d\pi(x,y))^{1/2}$, where $\Pi(\mu,\nu)$  the  set  of  all  probability measures $\Pi$ over the product set $\mathbb{R}\times\mathbb{R}$ with marginals $\mu$ and $\nu$.
The squared Wasserstein-2 distance between two multivariate Gaussian distributions $\mathbb{N}( m_{1},\Sigma_{1})$ and $\mathbb{N}( m_{2},\Sigma_{2})$ with diagonal covariances is : $\Vert m_1-m_2\Vert_2^2 +\Vert\Sigma_1^{1/2}-\Sigma_2^{1/2}\Vert_{F}^2$, where $\Vert \cdot \Vert_{F}$ represents the Frobenius norm.

\paragraph{\textbf{Proposition 1}} \textit{For a given DistGP $F$ and a Gaussian measure $\mu \sim \mathcal{N}(m_{1},\Sigma_{1}) $ to be the centre of an annulus $B(x) = \{ \nu\sim \mathcal{N}(m_{2},\Sigma_{2}|$ $ 0.125 \leq \frac{W_{2}(\mu,\nu)}{l^{2}} \leq 1.0$ and choosing any $\nu$ inside the ball we have the following Lipschitz bounds: $ W_{2}(F(\mu),F(\nu)) \leq L W_{2}(\mu,\nu)$, where  $L = (\frac{4\sigma^{2}}{l})^{2} \left[ \Vert K_{Z}^{-1}m \Vert_{2}^{2} + \Vert K_{Z}^{-1} \left( K_{z} - S\right)K_{Z}^{-1} \Vert_{2} \right]$ and  $l,\sigma^{2}$ are the lengthscales and variance of the kernel  .}

Proof is given in Sec.~\ref{subsec:proofs}. This theoretical result shows that the \emph{DistGP activation functions} have Lipschitz constants with respect to the Wasserstein-2 metric in both output and input domain. It is of vital importance to ensure the hidden layers $F_{l}^{pre}$ preserve the distance in Wasserstein-2 space in relation to the one at $F_{l-1}^{post}$, especially taking into consideration that we apply convolutional affine operators (Eq.~\ref{eq:conv_mu},~\ref{eq:conv_var}), which could break the smoothness of DistGP activations. This will ensure that the distance between previously identified outliers and inliers will stay constant.

\paragraph{\textbf{Proposition 2}} \textit{We consider the affine operator $A \in \R^{C,1}$ operating in the space of multivariate Gaussian measures of size C. Consider two distributions $\mu \sim \mathcal{N}(m_{1}, \sigma^{2}_{1})$ and $\nu \sim \mathcal{N}(m_{2}, \sigma^{2}_{2})$, which can be thought of as elements of a hidden layer patch, then for the affine operator function $f(\mu) = \mathbb{N}(m_{1} A, \sigma_{1}^{2}A^{2})$ we have the following Lipschitz bound:}
$
W_{2}(f(\mu), f(\nu)) \le L W_{2}(\mu, \nu))
$\textit{, where $L = \sqrt{C}\Vert W \Vert_2^{2}$.}

Proof is given in Sec.~\ref{subsec:proofs}. We denote the l-th layer weight matrix, computing the c-th channel by column matrix $A_{l,c}$. We can impose the Lipschitz condition to Eq.~\ref{eq:conv_mu},~\ref{eq:conv_var} by having constrained weight matrices with elements of the form $A_{l,c} = \frac{A_{l,1}}{C^{\frac{1}{4}}\sqrt{\sum_{c=1}^{C}W_{l,c}^{2}}}$. 






\subsection{Proving Lipschitz bounds in a DistGP layer}
\label{subsec:proofs}

We here prove Propositions 1 and 2 of Sec.~\ref{sec:conv_dist_gp}. 

\paragraph{\textbf{Definition}} The Wasserstein-2 distance between two multivariate Gaussian distributions $\mathbb{N}( m_{1},\Sigma_{1})$ and $\mathbb{N}( m_{2},\Sigma_{2})$ with diagonal covariances is : $\Vert m_1-m_2\Vert_2^2 +\Vert\sigma_1^{1/2}-\sigma_2^{1/2}\Vert_{F}^2$, where $\Vert \cdot \Vert_{F}$ represents the Frobenius norm.

\paragraph{\textbf{Lemmas on p-norms}} We have the following relations between norms :
$\Vert x \Vert_{2} \leq \Vert x \Vert_{1} $ and $\Vert x \Vert_{1} \leq \sqrt{D} \Vert x \Vert_{2}$. Will be used for the proof of Proposition 2.

\paragraph{\textbf{Proof of Proposition 1}}

Throughout this subsection we shall refer to the first two moments of a Gaussian measure by $m(\cdot)$, $v(\cdot)$. Explicitly writing the Wasserstein-2 distances of the inequality we get: 
\begin{equation}
    |m(F(\mu)) - m(F(\nu))|^{2} + |v(F(\mu)) - v(F(\nu)|^{2} \leq L |m_{1} - m_{2}|^{2} +  |\Sigma_{1} - \Sigma_{2}|^{2}
\end{equation}
 We focus on the mean part and applying Cauchy–Schwarz we get the following inequality:
 \begin{equation}
     | \left[ K_{\mu,Z} - K_{\nu,Z} \right] K_{Z}^{-1}m |^{2} \leq \Vert K_{\mu,Z} - K_{\nu,Z} \Vert_{2}^{2} \Vert K_{Z}^{-1}m \Vert_{2}^{2}     
 \end{equation}
To simplify the problem and without loss of generality we consider $U_{z}$ to be a sufficient statistic for the set of inducing points $Z$. Expanding the first term of the r.h.s. we get: 
\begin{equation}
    \left[ \sigma^{2} \exp{\frac{-W_{2}(\mu,U_{z})}{l^{2}}} - \sigma^{2} \exp{\frac{-W_{2}(\nu, U_{z}) } {l^{2}}} \right]    
\end{equation}
We assume $\nu = \mu + h$, where $h \sim \mathcal{N}(|m_{1}-m_{2}|,|\Sigma_{1} - \Sigma_{2}|)$ and  $\mu$ is a high density point in the data manifold, hence $W_{2} (\mu - U_{z}) = 0$. We denote $m(h)^{2}+var(h)^{2}=\lambda$. Considering the general equality $\log(x-y) = \log(x) + \log(y) + \log(\frac{1}{y} - \frac{1}{x})$ and applying it to our case we obtain: 
\begin{gather}
    \log |m(F(\mu)) - m(F(\nu))|^{2} \leq \log\left[ \sigma^{2} -\sigma^{2} \exp\frac{{-\lambda}}{l^{2}} \right]^{2} \\ 
    = 2\log \sigma^{2} - 2 \frac{\lambda}{l^{2}} + 2\log \left[ \exp{\frac{\lambda}{l^{2}}} -1 \right] \leq 2 \log \left[ \sigma^{2} \exp{\frac{\lambda}{l^{2}}}  \right]     
\end{gather}
We have the general inequality $\exp{x} \leq 1+x+x^{2}$ for $x\leq 1.79$, which for $0\leq x \leq 1$ can be modified as $\exp{x} \leq 1+2x$. Applying this new inequality we have the following: 
\begin{equation}
|m(F(\mu)) - m(F(\nu))|^{2} \leq \left[ \sigma^{2} + 2\sigma^{2}\frac{\lambda}{l^{2}}\right]^{2} = \sigma^{4} + \sigma^{4} \frac{\lambda}{l^{2}} + 4 \sigma^{4} \frac{(\lambda)^{2}}{l^{4}} \leq 16 \sigma^{4}\frac{\lambda}{l^{2}}    
\end{equation}
where the last inequality follows from the annulus constrains. We now move to the variance components of the Lipschitz bound. We can notice that:
\begin{equation}
    | v(F(\mu))^{\frac{1}{2}} - v(F(\nu))^{\frac{1}{2}} |^{2} \leq | v(F(\mu))^{\frac{1}{2}} - v(F(\nu))^{\frac{1}{2}} || v(F(\mu))^{\frac{1}{2}} + v(F(\nu))^{\frac{1}{2}} | = | v(F(\mu)) - v(F(\nu))|
\end{equation}
which after applying Cauchy–Schwarz results in an upper bound of the following form $\Vert K_{\mu,U_{z}} - K_{\nu,U_{z}}\Vert_{2}^{2} \Vert K_{U_{z}}^{-1}(K_{U_{z}-S})K_{U_{z}}^{-1}\Vert_{2}$. Using that $\Vert K_{\mu,U_{z}} - K_{\nu,U_{z}} \Vert_{2}^{2} \leq  \frac{16\sigma^{4} \lambda}{l^{2}}$ we obtain that:
\begin{equation}
    |v(F(\mu)) - v(F(\nu))| \leq \frac{16\sigma^{4}\lambda}{l^{2}} \Vert K_{U_{z}}^{-1}(K_{U_{z}}-S)K_{U_{z}}^{-1}\Vert_{2}   
\end{equation}
Now taking into consideration both the upper bounds on the mean and variance components we arrive at the desired Lipschitz constant.

\paragraph{\textbf{Proof of Proposition 2}}


Using the definition for Wasserstein-2 distances and taking the l.h.s of the inequality, we obtain: 
\begin{equation}
    \Vert m_{1}A-m_{2}A\Vert_2^2 +\Vert(\sigma_{1}^{2}A^{2})^{1/2}-(\sigma_{2}^{2}A^{2})^{1/2}\Vert_{F}^2
\end{equation}
which after rearranging terms and noticing that inside the Frobenius norm we have scalars, becomes: 
\begin{equation}
    \Vert (m_{1}-m_{2})A\Vert_2^2+ [\sigma_{1}^{2}A^{2})^{1/2}-(\sigma_{2}^{2}A^{2})^{1/2}]^2    
\end{equation}
We can now apply the Cauchy–Schwarz inequality for the part involving means and multiplying the right hand side with $\sqrt{C}$, which represents the number of channels, we get: 
\begin{equation}
    \Vert (m_{1}-m_{2})A\Vert_2^2 + [\sigma_{1}^{2}A^{2})^{1/2}-(\sigma_{2}^{2}A^{2})^{1/2}]^2 \leq \Vert m_{1}-m_{2} \Vert_2^{2} \sqrt{C}\Vert A \Vert_2^{2} + \sqrt{C} [\sigma_{1}^{2}A^{2})^{1/2}-(\sigma_{2}^{2}A^{2})^{1/2}]^2    
\end{equation}
We can notice that the Lipschitz constant for the component involving mean terms is $\sqrt{C}\Vert A \Vert_2^{2}$. Hence, we try to prove that the same L is also available for the variance terms component. 
Hence, we have the following if and only if statement:
\begin{equation}
    L=\sqrt{C}\Vert A \Vert_2^{2}  \leftrightarrow \sqrt{C} [\sigma_{1}^{2}A^{2})^{1/2}-(\sigma_{2}^{2}A^{2})^{1/2}]^2 \leq [\sigma_{1}-\sigma_{2}]^{2}  \sqrt{C}\Vert A \Vert_2^{2}    
\end{equation}
By virtue of Cauchy–Schwarz we have the following inequality $\sqrt{C}[\sigma_{1}A-\sigma_{2}A]^{2} \leq [\sigma_{1}-\sigma_{2}]^{2}  \sqrt{C}\Vert A \Vert_2^{2}$. Hence the aforementioned if and only if statement will hold if we prove that: 
\begin{equation}
    \sqrt{C}\left[(\sigma_{1}^{2}A^{2})^{\frac{1}{2}} - (\sigma_{2}^{2}A^{2})^{\frac{1}{2}}\right]^{2} \leq \sqrt{C}\left[\sigma_{1}A-\sigma_{2}A\right]^{2}    
\end{equation}
which after expressing in terms of norms becomes:
\begin{gather}
    \sqrt{C}\left[ \Vert\sigma_{1}A\Vert_{2} - \Vert\sigma_{2}A\Vert_{2}\right]^{2} \leq \sqrt{C}\left[ \Vert\sigma_{1}A\Vert_{1} - \Vert\sigma_{2}A\Vert_{1} \right]^{2} \\
    \sqrt{C}\left[\Vert\sigma_{1}A\Vert_{2}^{2}+ \Vert\sigma_{2}A\Vert_{2}^{2} -  2\Vert\sigma_{1}A\Vert_{2}\Vert\sigma_{2}A\Vert_{2}\right] \leq \sqrt{C}\left[\Vert\sigma_{1}A\Vert_{1}^{2}+\Vert\sigma_{2}A\Vert_{1}^{2}-2\Vert\sigma_{1}A\Vert_{1}\Vert\sigma_{2}A\Vert_{1}\right]
\end{gather}
This inequality holds by applying the p-norm lemmas, thereby the if and only if statement is satisfied. Consequently, the Lipschitz constant is $\sqrt{C}\Vert A \Vert_2^{2}$.

\section{DistGP-based Segmentation Network \& OOD detection}

\begin{figure}[!htb]
  \includegraphics[width=\linewidth, height=10cm]{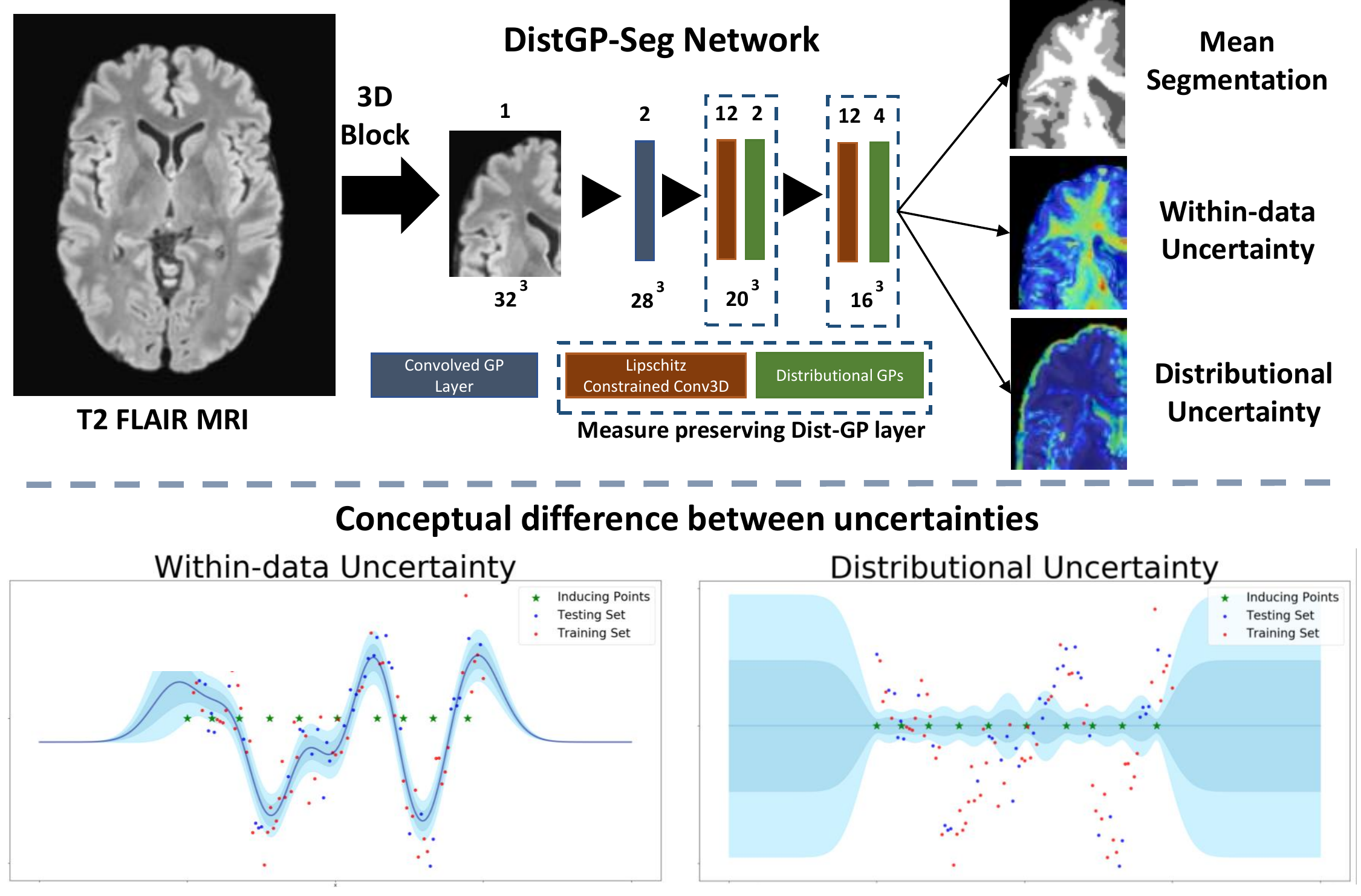}
  \caption{\textbf{Top}: Schematic of proposed DistGP activated segmentation net. Above and below each layer we show the number of channels and their dimension respectively. \textbf{Bottom}: Visual depiction of the two uncertainties in DistGP after fitting a toy regression manifold. Distributional uncertainty increases outside the manifold of training data and is therefore useful for OOD detection.}
  \label{fig:schematic_seg_net_and_uncertainties}
\end{figure}

The above introduced modules in Sec.~\ref{sec:conv_dist_gp} can be used to construct a convolutional network that benefits from properties of DistGP. Specifically, we construct a 3D network for segmenting volumetric medical images, which is depicted in Figure \ref{fig:schematic_seg_net_and_uncertainties} (top). It consists of a convolved GP layer, followed by two measure-preserving DistGP layers. Each hidden layer uses filters of size $5\!\times\!5\!\times\!5$. To increase the model's receptive field, in the second layer we use convolution dilated by 2. We use 250 inducing points and 2 channels for the DistGP "activation functions". The affine operators project the stochastic patches into a 12 dimensional space. The size of the network is limited by computational requirements for GP-based layers, which is an active research area. Like regular convolutional nets, this model can process input of arbitrary size but GPU memory requirement increases with input size. We here provide input of size $32^{3}$  to the model, which then segments the central $16^{3}$ voxels. To segment a whole scan we divide it into tiles and stitch together the segmentations.

While prediction uncertainty can be computed for standard neural networks by using the softmax probability, these uncertainty estimates are often overconfident \cite{guo2017calibration,mcclure2019knowing}, and are less reliable than those obtained by Bayesian networks \cite{nair2020exploring}. For our model we decompose the model uncertainty into two components by splitting the last DistGP layer in two parts:
$h(\cdot) =  \mathcal{N}(h|0, K_{nn} - K_{nm}K_{mm}^{-1}K_{mn})$  and $
g(\cdot) = \mathcal{N}(g|K_{nm}K_{mm}^{-1}m, K_{nm}K_{mm}^{-1} S K_{mm}^{-1}  K_{mn})$. The $h(\cdot)$ variance captures the shift from within to outside the data manifold and will be denoted as \emph{distributional uncertainty}. The variance $g(\cdot)$ is termed here as \emph{within-data uncertainty} and encapsulates uncertainty present inside the data manifold. A visual depiction of the two is provided in Figure \ref{fig:schematic_seg_net_and_uncertainties} (bottom).

\section{Evaluation on Brain MRI}

In this section we evaluate our method alongside recent OOD models \cite{van2020simple,franchi2020one,padhy2020revisiting}, assessing their capabilities to reach segmentation performance comparable to well-established deterministic models and whether they can accurately detect outliers.

\subsection{Data and pre-processing}

For evaluation we use publicly available datasets:

1) Brain MRI scans from the UKBB study \cite{alfaro2018image}, which contains scans from nearly 15,000 subjects. We selected for training and evaluation the bottom 10$\%$ percentile in terms of white matter hypointensities with an equal split between training and testing. All subjects have been confirmed to be normal by radiological assessment. Segmentation of brain tissue (CSF,GM,WM) has been obtained with SPM12.

2) MRI scans of 285 patients with gliomas from BraTS 2017 \cite{bakas2017advancing}. All classes are fused into a \emph{tumor} class, which we will use to quantify OOD detection performance.

In what follows, we use only the FLAIR sequence to perform the brain tissue segmentation task and OOD detection of tumors, as this MRI sequence is available for both UKBB and BraTS. All FLAIR images are pre-processed with skull-stripping, N4 bias correction, rigid registration to MNI152 space and histogram matching between UKBB and BraTS. Finally, we normalize intensities of each scan via linear scaling of its minimum and maximum intensities to the [-1,1] range.

\subsection{Brain tissue segmentation on normal MRI scans}

\begin{table}[h!]
\begin{center}
\caption{Performance on UK Biobank in terms of Dice scores per tissue.}
\label{tab:results}
 \begin{tabular}{c c c c c} 
 \toprule
 Model & Hidden Layers &  DICE CSF & DICE GM & DICE WM \\ [0.5ex] 
 \midrule
 OVA-DM \cite{padhy2020revisiting}    & 3         &   0.72       & 0.79       & 0.77  \\
 OVNNI \cite{franchi2020one}    & 3         &   0.66       & 0.77       & 0.73   \\
 DUQ \cite{van2020simple}       & 3         & 0.745      & 0.825    & 0.781 \\ 
 DistGP-Seg (ours) & 3 & 0.829 & 0.823 & 0.867 \\
\midrule
 U-Net     & 3 scales  & 0.85 & 0.89 & 0.86 \\ 
\bottomrule
\end{tabular}
\end{center}
\end{table}

\paragraph{\textbf{Task:}} We train and test our model on segmentation of brain tissue of healthy UKBB subjects. This corresponds to the within-data manifold in our setup.

\paragraph{\textbf{Baselines:}} 
We compare our model with recent Bayesian approaches for enabling task-specific models (such as image segmentation) to perform uncertainty-based OOD detection \cite{van2020simple,franchi2020one,padhy2020revisiting}. For fair comparison, we use these methods in an architecture similar to ours (Figure ~\ref{fig:schematic_seg_net_and_uncertainties}), except that each layer is replaced by standard convolutional layer, each with 256 channels, LeakyRelu activations, and dilation rates as in ours. We also compare these Bayesian methods with a well-established deterministic baseline, a U-Net with 3 scales (down/up-sampling) and 2 convolution layers per scale in encoder and 2 in decoder (total 12 layers).

\paragraph{\textbf{Results:}}
Table \ref{tab:results} shows that DistGP-Seg surpasses other Bayesian methods with respect to Dice score for all tissue classes. Our method approaches the performance of the deterministic U-Net, which has a much larger architecture and receptive field. We emphasize this has not been previously achieved with GP-based architectures, as their size (e.g. number of layers) is limited due to computational requirements. This supports the potential of DistGP, which is bound to be further unlocked by advances in scaling GP-based models.

\subsection{Outlier detection in MRI scans with tumors}

\begin{table}[h!]
\begin{center}
\caption{Performance comparison of Dice for detecting outliers on BraTS for different thresholds obtained from UKBB.}
\label{tab:ood_results}
 \begin{tabular}{c c c c c} 
 \toprule
Model  & \multicolumn{1}{p{2cm}}{\centering DICE \\ FPR=0.1 }  & \multicolumn{1}{p{2cm}}{\centering DICE \\ FPR=0.5 } & \multicolumn{1}{p{2cm}}{\centering DICE \\ FPR=1.0 } & \multicolumn{1}{p{2cm}}{\centering DICE \\ FPR=5.0 } \\ [0.5ex] 
 \midrule
 OVA-DM \cite{padhy2020revisiting}       &    0.382      &    0.428    &  0.457  & 0.410 \\
 OVNNI \cite{franchi2020one}         &    $\leq 0.001 $      & $\leq 0.001 $ & $\leq 0.001 $   & $\leq 0.001 $ \\
 DUQ \cite{van2020simple}    & 0.068     &  0.121    &   0.169    & 0.182 \\
 DistGP-Seg (ours) & 0.512 & 0.571 & 0.532 & 0.489 \\
\midrule
 VAE-LG \cite{chen2019unsupervised}          &  0.259      &  0.407      &  0.448  & 0.303 \\
 AAE-LG \cite{chen2019unsupervised}            &   0.220     &   0.395     &  0.418  & 0.302 \\
\bottomrule
\end{tabular}
\end{center}
\end{table}

\paragraph{\textbf{Task:}} The previous task of brain tissue segmentation on UKBB serves as a proxy task for learning normative patterns with our network. Here, we apply this pre-trained network on BRATS scans with tumors. We expect the region surrounding the tumor and other related pathologies, such as squeezed brain parts or shifted ventricles, to be highlighted with higher distributional uncertainty, which is the OOD measure for the Bayesian deep learning models. To evaluate quality of OOD detection at a pixel level, we follow the procedure in \cite{chen2019unsupervised}, for example to get the 5.0$\%$ False Positive Ratio threshold value we compute the 95$\%$ percentile of distributional variance on the testing set of UKBB, taking into consideration that there is no outlier tissue there. Subsequently, using this value we threshold the distributional variance heatmaps on BraTS, with tissue having a value above the threshold being flagged as an outlier. We then quantify the overlap of the pixels detected as outliers (over the threshold) with the ground-truth tumor labels by computing the Dice score between them. 

\begin{figure*}[!htb]
    \centering
    \includegraphics[width=0.95\linewidth]{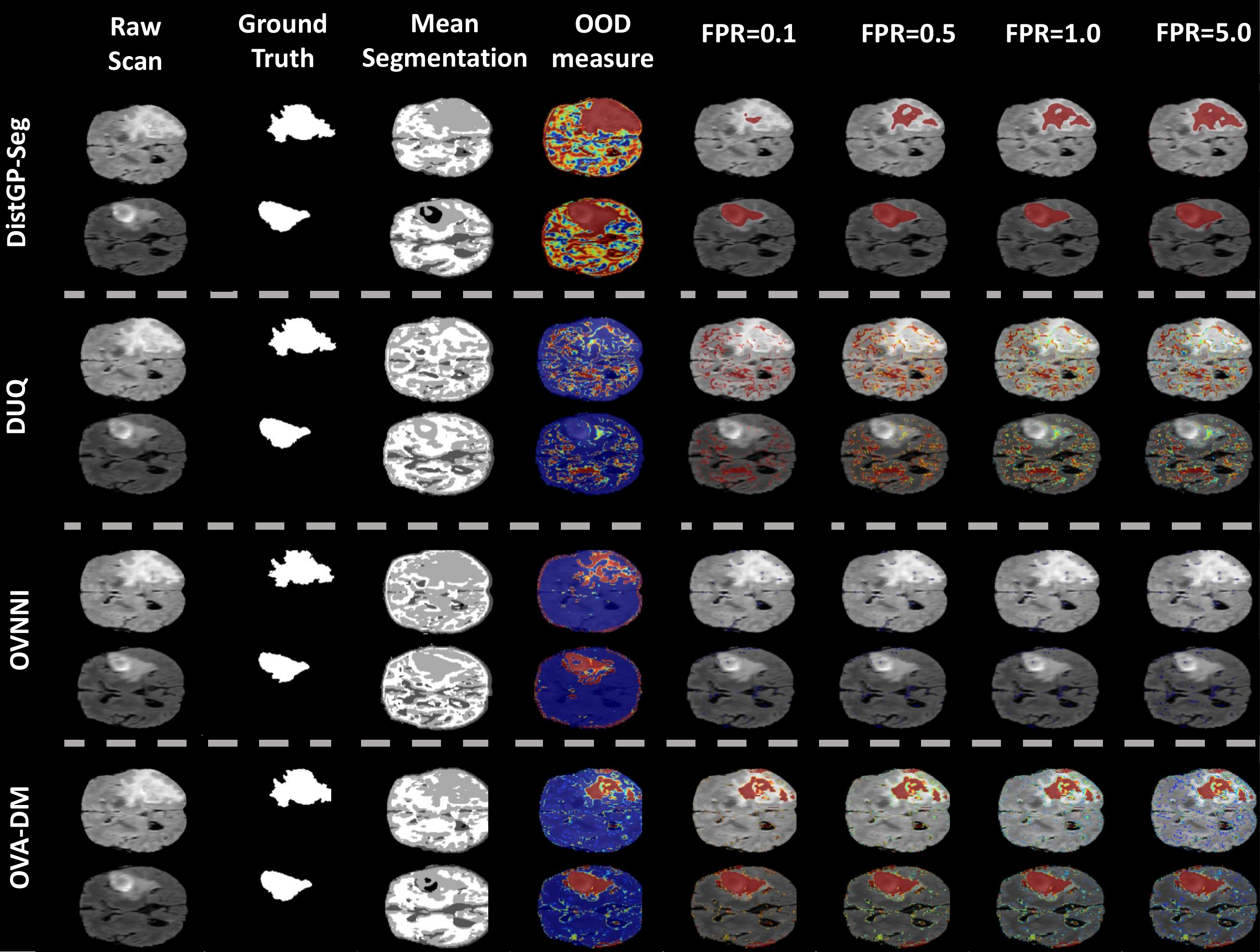}
    \caption{Comparison between models in terms of voxel-level outlier detection of tumors on BRATS scans. Mean segmentation represents the hard segmentation of brain tissues. OOD measure is the quantification of uncertainty for each model, using their own procedure. Higher values translate to appartenance to outlier status, whereas for OVNNI it is the converse.}
    \label{fig:tumor_class_comparison_brats}  
\end{figure*}
\clearpage

\paragraph{\textbf{Results:}} Table \ref{tab:ood_results} shows the results from our experiments with DistGP and compared Bayesian deep learning baselines. We also provide performance of reconstruction-based OOD detection models as reported in \cite{chen2019unsupervised} for similar experimental setup. DistGP-Seg surpasses its Bayesian deep learning counterparts, as well as reconstructed-based models. In Figure \ref{fig:tumor_class_comparison_brats} we provide representative results from the methods we implemented for qualitative assessment. Moreover, although BRATS does not provide labels for WM/GM/CSF tissues hence we cannot quantify how well these tissues are segmented, visual assessment shows our method compares favorably to compared counterparts.

\section{Discussion}

We have introduced a novel Bayesian convolutional layer with Lipschitz continuity that is capable of reliably propagating uncertainty . General criticism surrounding deep and convolutional GP involves the issue of under-performance compared to other Bayesian deep learning techniques, and especially compared to deterministic networks. Our experiments demonstrate that our 3-layers model, size limited due to computational cost, is capable of approaching the performance of a U-Net, an architecture with a much larger receptive field. Further advances in computational efficient GP-based models, an active area of research, will enable our model to scale further and unlock its full potential. Importantly, we showed that our DistGP-Seg network offers better uncertainty estimates for OOD detection than the state-of-the-art Bayesian approaches, and also surpasses recent unsupervised reconstruction-based deep learning models for identifying outliers corresponding to pathology on brain scans. Our results indicate that OOD methods that do not take into account distances in latent space, such as OVNNI, tend to fail in detecting outliers, whereas OVA-DM and DUQ that make predictions based on distances in the last layer perform better. Our model utilises distances at every hidden layer, thus allowing the notion of outlier to evolve gradually through the depth of our network. This difference can be noticed in the smoothness of OOD measure for our model in comparison to other methods in Figure \ref{fig:tumor_class_comparison_brats}. A drawback of our study resides in the small architecture used. Extending our ``measure preserving DistGP'' module to larger architectures such as U-Net for segmentation or modern CNNs for whole-image prediction tasks remains a prospective research avenue fuelled by advances in scalability of SGP. In conclusion, our work shows that incorporating DistGP in convolutional architectures provides both competitive performance and reliable uncertainty quantification in medical image analysis, opening up a new direction of research.

\bibliography{ms}
\bibliographystyle{abbrvnat}

\end{document}


\section{Supplementary Material}

\subsection{Additional Figures for Detecting Outliers on Brain Scans}
    
\subsubsection{Distributional GP-activated Segmentation}

\begin{figure}[!t]
    \centering
    \includegraphics[width=\linewidth]{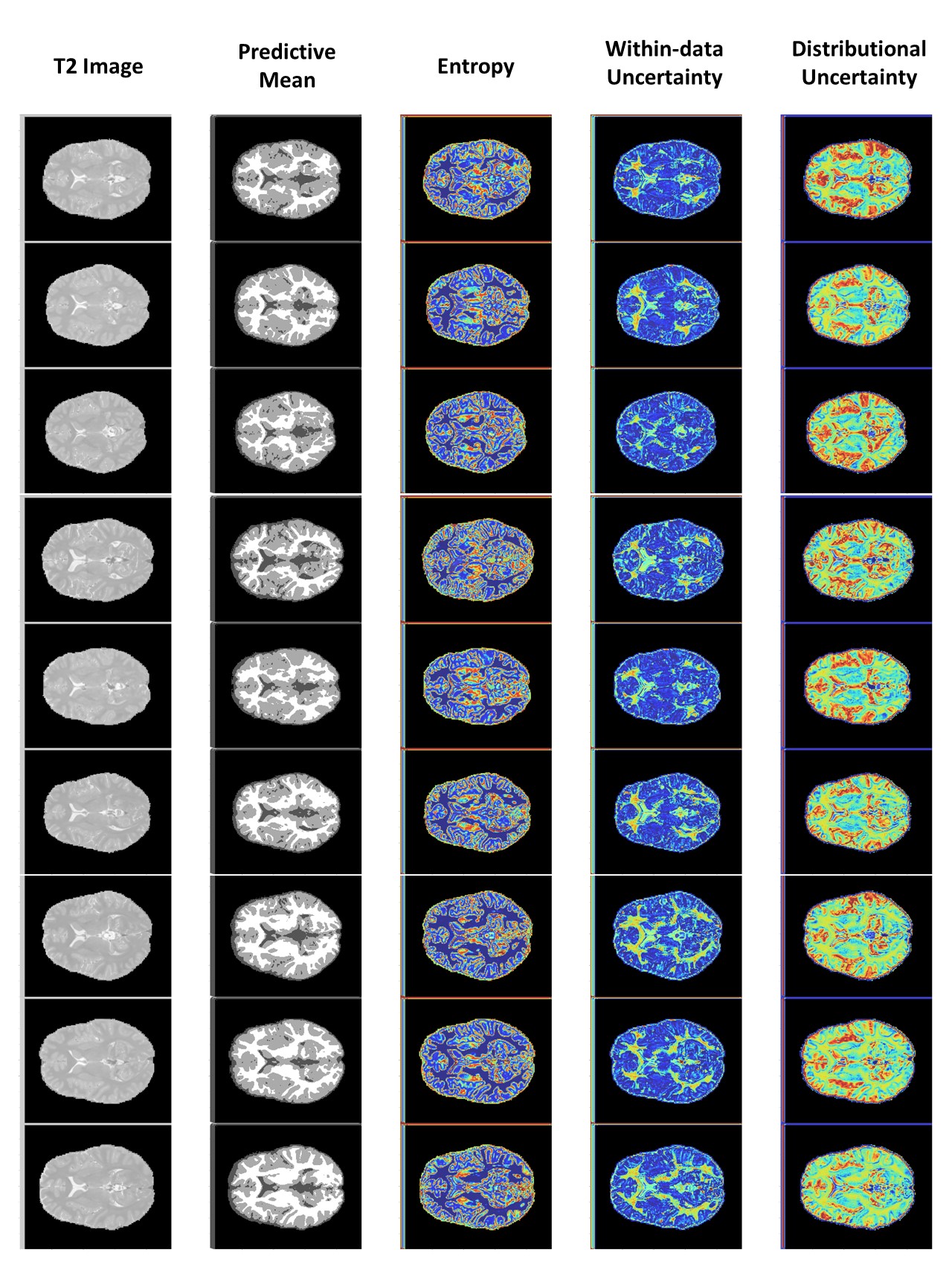}
    \caption{Segmentations on 9 healthy subjects from CamCan using DistributionalGP activated Baseline DeepMedic. Epistemic differential entropy is a measure of within the data manifold uncertainty, whereas distributional differential entropy detects outliers. Higher values indicated more uncertainty.Each column's colormap for different subjects is within the same interval.}
    \label{}  
\end{figure}

\begin{figure}[!t]
    \centering
    \includegraphics[width=\linewidth]{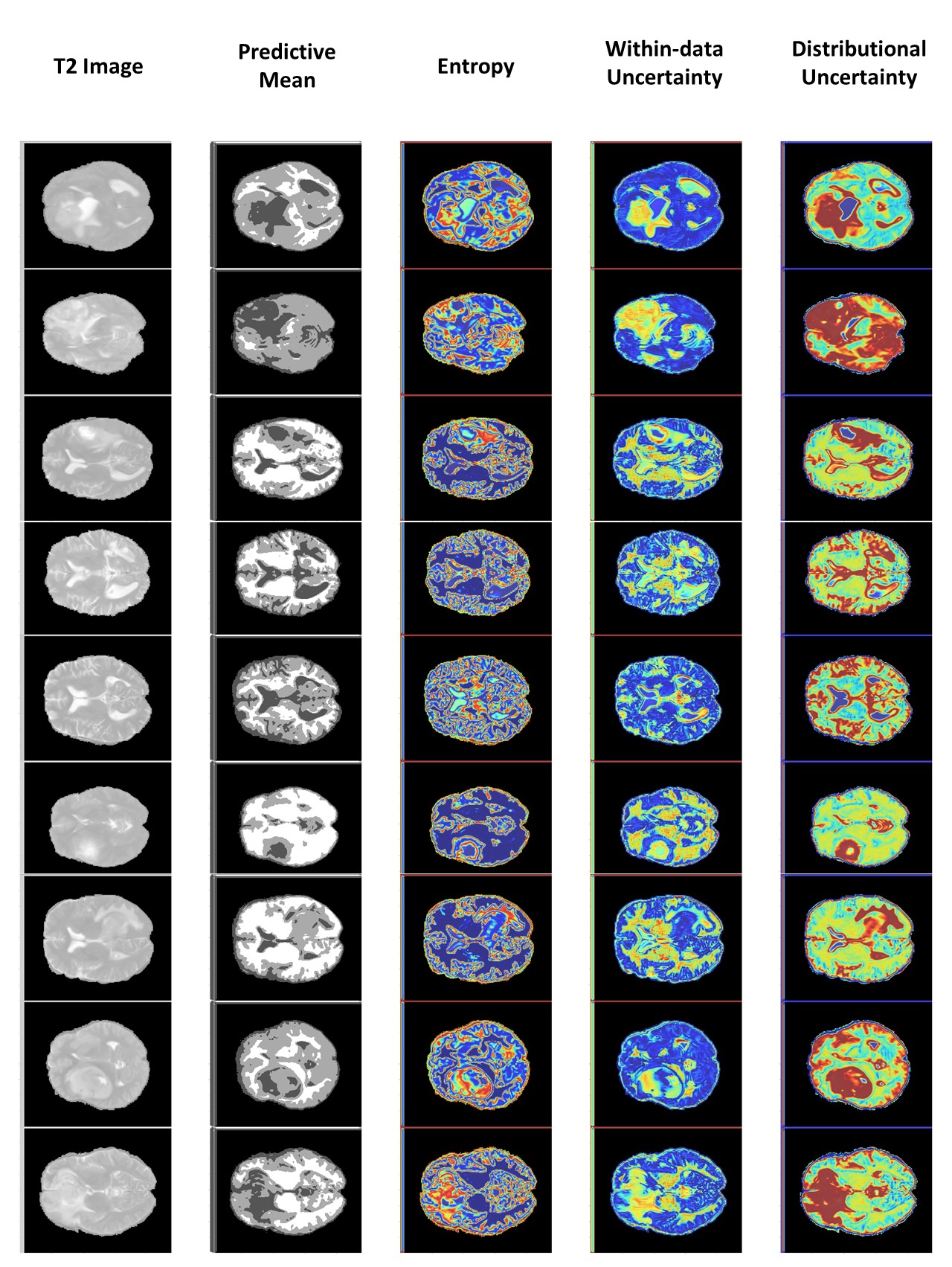}
    \caption{Segmentations on 9 subjects with tumors from BRATS using DistributionalGP activated Baseline DeepMedic. Epistemic differential entropy is a measure of within the data manifold uncertainty, whereas distributional differential entropy detects outliers. Higher values indicated more uncertainty.Each column's colormap for different subjects is within the same interval.}
    \label{}  
\end{figure}

\begin{figure}[!t]
    \centering
    \includegraphics[width=\linewidth]{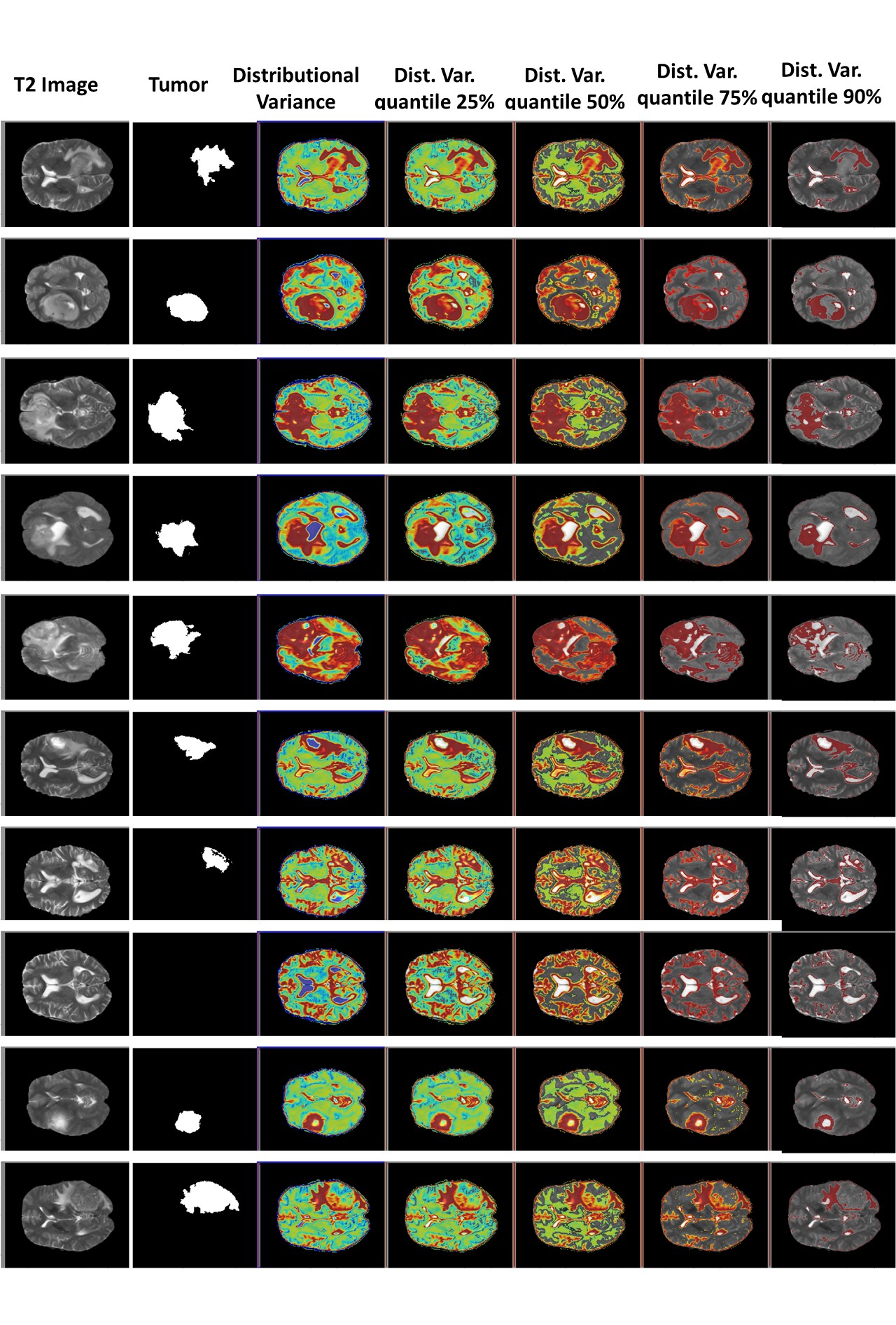}
    \caption{Overlap between ground-truth segmentation of the tumor and quantile thresholded distributional variance based segmentations of outlier. Results are from Distributional GP activated Baseline DeepMedic.}
    \label{fig:tumor_variance_filtering}  
\end{figure}

\subsubsection{Deterministic Uncertainty Quantification}

\begin{figure}[!t]
    \centering
    \includegraphics[width=\linewidth]{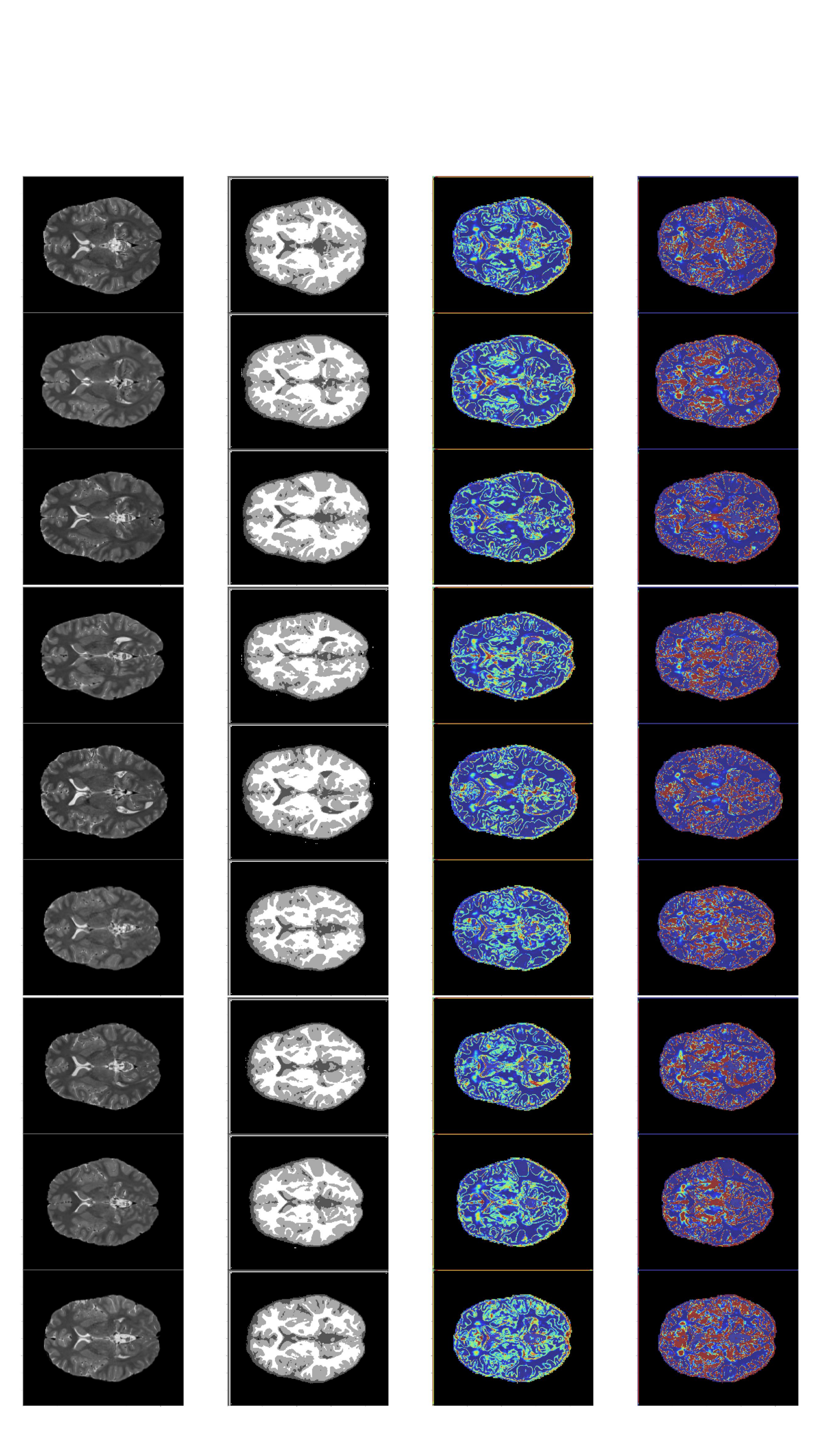}
    \caption{Segmentations on 9 healthy subjects from CamCan using DUQ. Distance is a measure of nearness of the respective voxel latent embedding with respect to the optimized centroids. Higher values indicated more uncertainty.Each column's colormap for different subjects is within the same interval.}
    \label{}  
\end{figure}

\begin{figure}[!t]
    \centering
    \includegraphics[width=\linewidth]{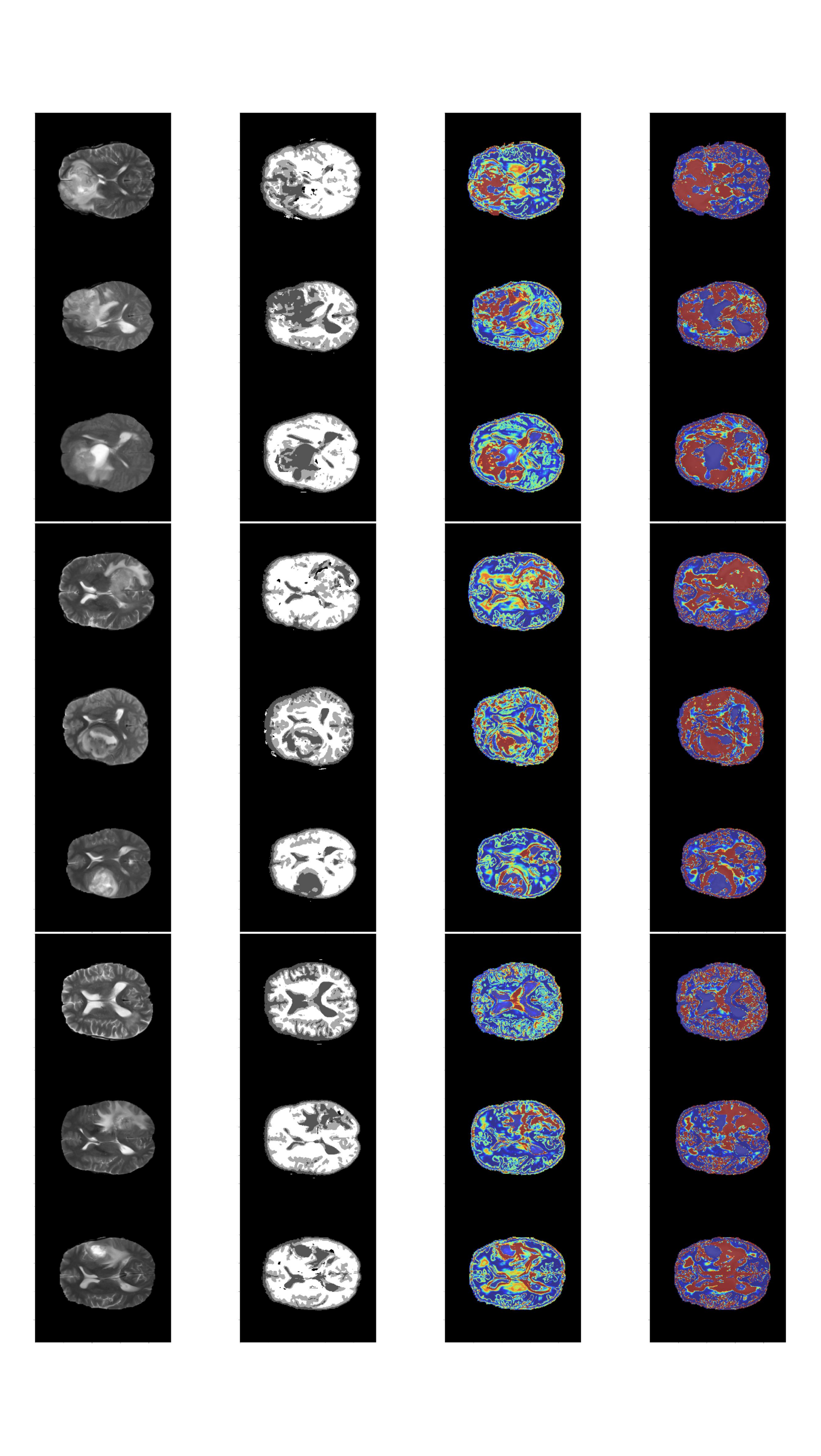}
    \caption{Segmentations on 9 subjects with tumors from BRATS using DUQ. Distance is a measure of nearness of the respective voxel latent embedding with respect to the optimized centroids. Higher values indicated more uncertainty.Each column's colormap for different subjects is within the same interval.}
    \label{}  
\end{figure}

\begin{figure}[!t]
    \centering
    \includegraphics[width=\linewidth]{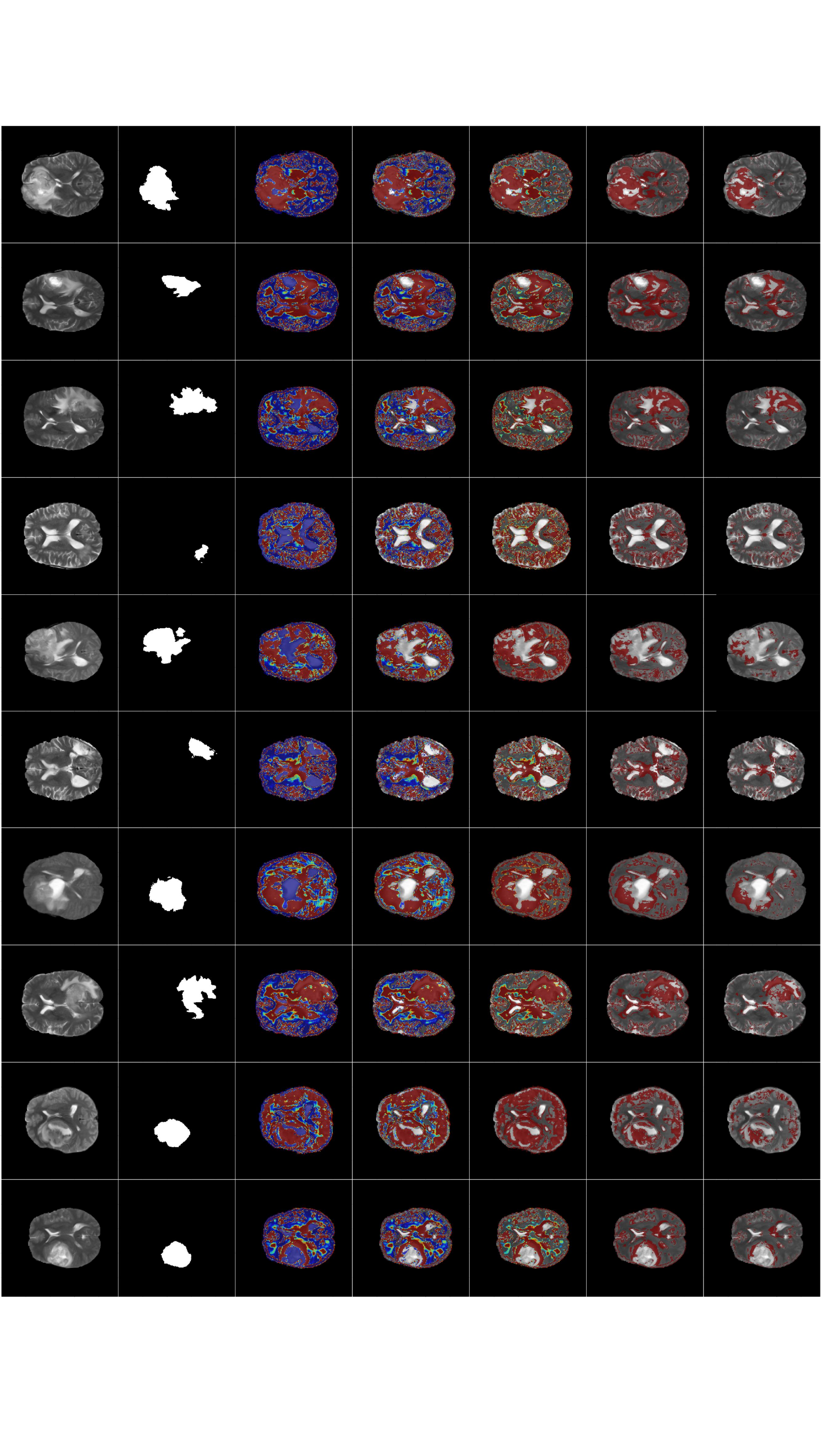}
    \caption{Overlap between ground-truth segmentation of the tumor and quantile thresholded distance based segmentations of outlier. Results are from DUQ.}
    \label{fig:tumor_variance_filtering}  
\end{figure}

\subsubsection{One versus All - DM}

\begin{figure}[!t]
    \centering
    \includegraphics[width=\linewidth]{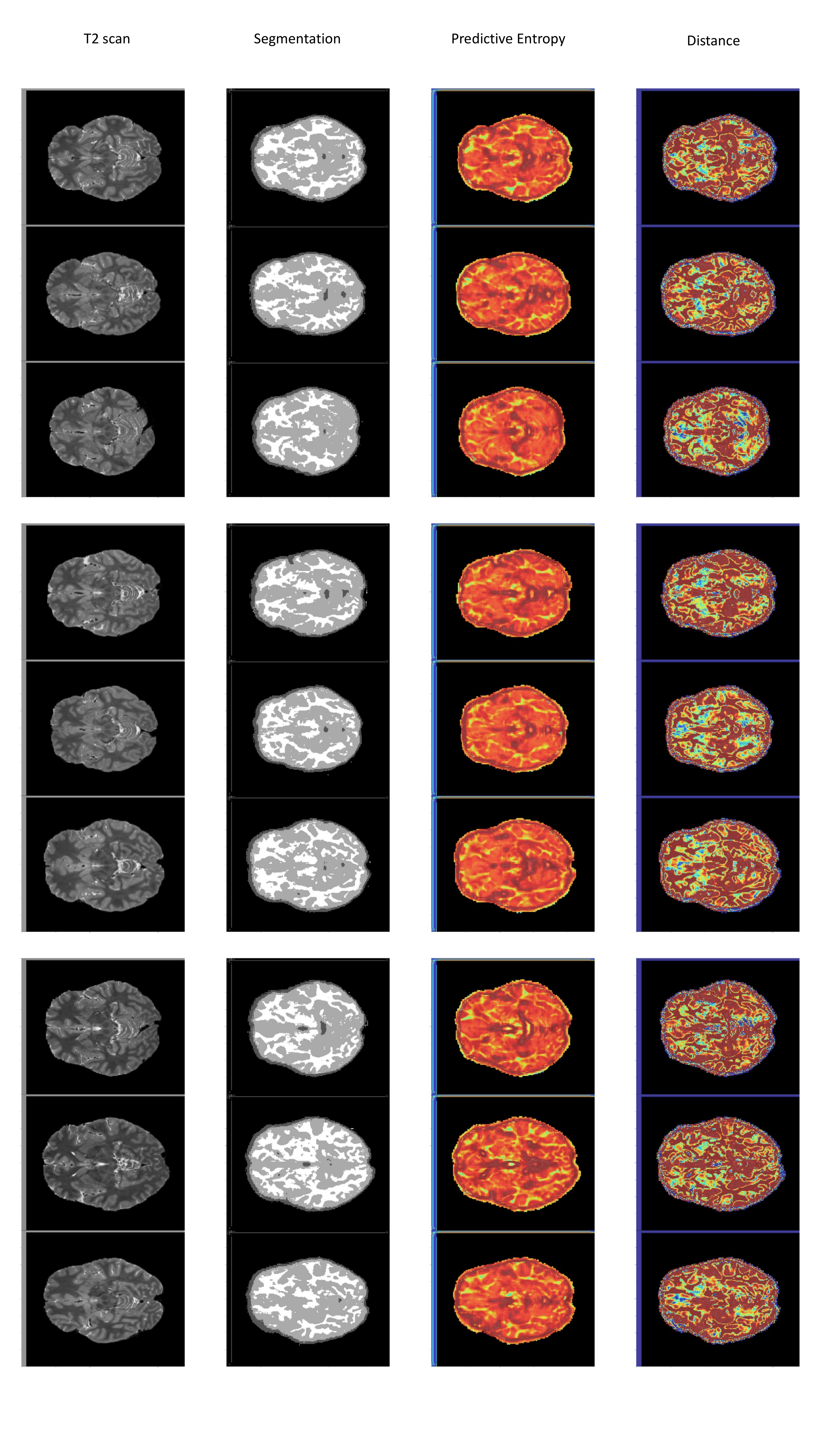}
    \caption{Segmentations on 9 healthy subjects from CamCan using OVA-DM. Distance is a measure of nearness of the respective voxel latent embedding with respect to the optimized centroids. Higher values indicated more uncertainty.Each column's colormap for different subjects is within the same interval.}
    \label{}  
\end{figure}

\begin{figure}[!t]
    \centering
    \includegraphics[width=\linewidth]{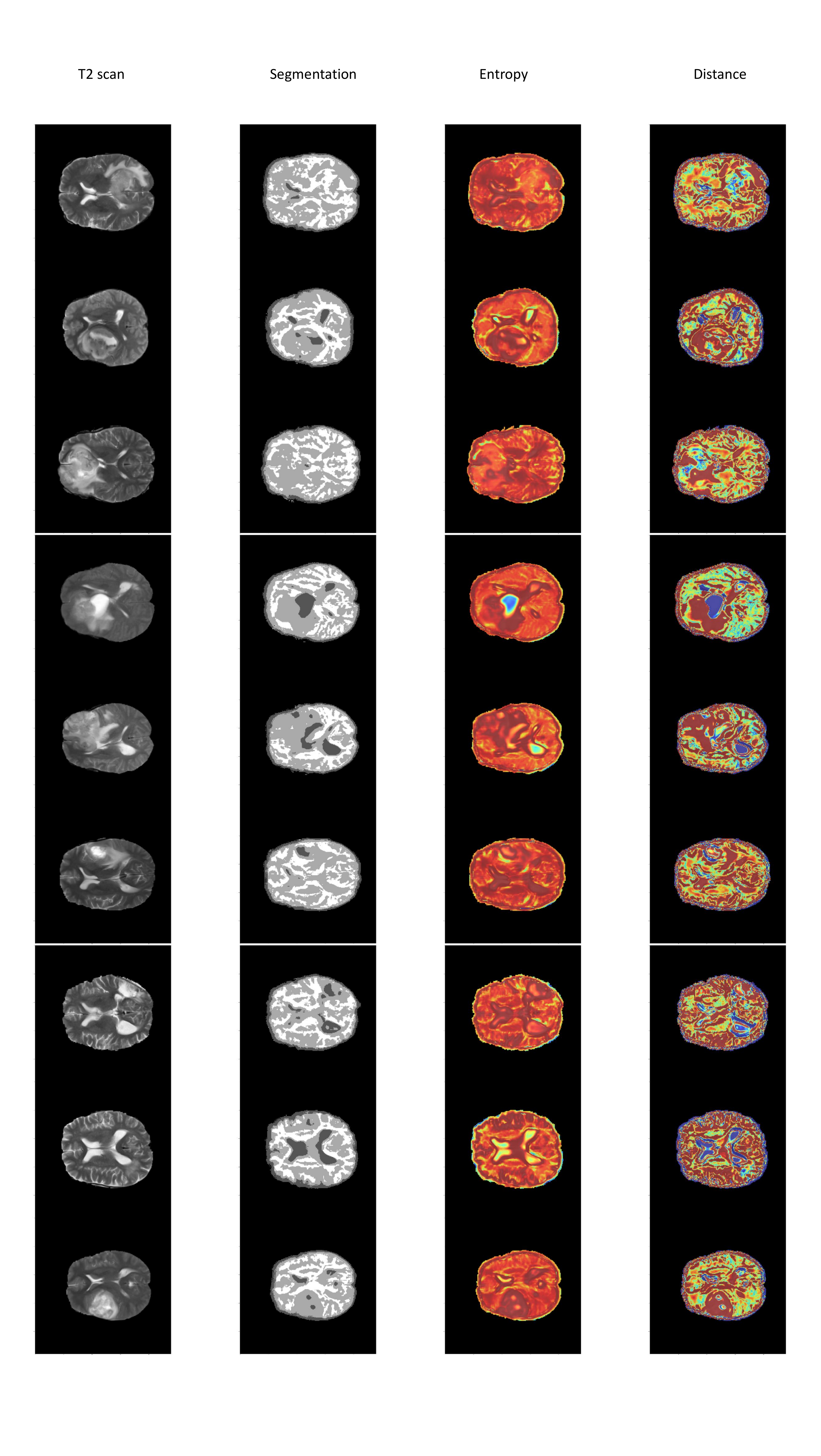}
    \caption{Segmentations on 9 subjects with tumors from BRATS using OVA-DM. Distance is a measure of nearness of the respective voxel latent embedding with respect to the optimized centroids. Higher values indicated more uncertainty.Each column's colormap for different subjects is within the same interval.}
    \label{}  
\end{figure}

\begin{figure}[!t]
    \centering
    \includegraphics[width=\linewidth]{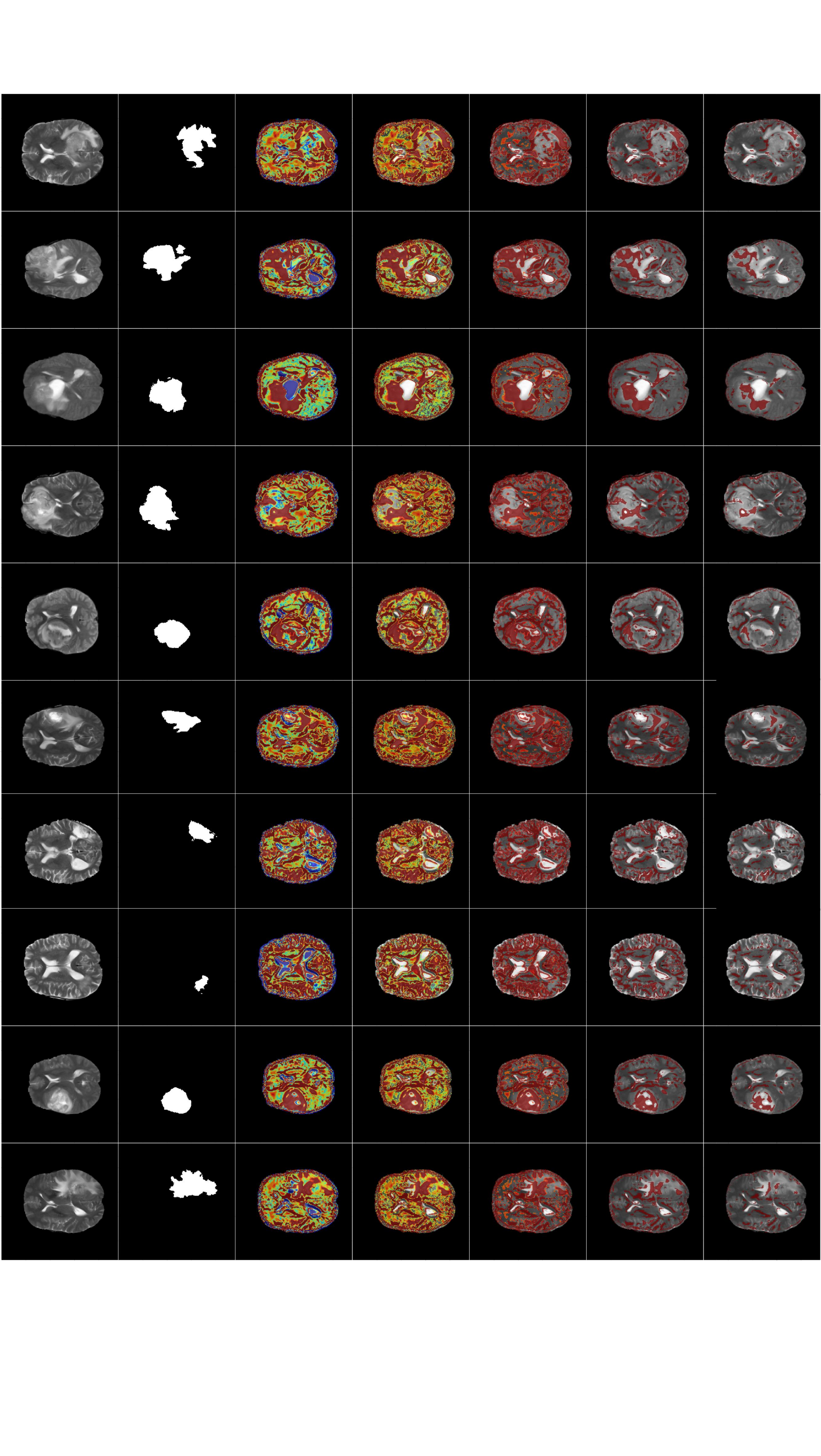}
    \caption{Overlap between ground-truth segmentation of the tumor and quantile thresholded distance based segmentations of outlier. Results are from OVA-DM.}
    \label{fig:tumor_variance_filtering}  
\end{figure}